\documentclass[letterpaper, 10 pt, conference]{ieeeconf}  

\IEEEoverridecommandlockouts                              

\usepackage{graphics} 
\usepackage{epsfig} 
\usepackage{mathptmx} 
\usepackage{times} 
\usepackage{amsmath} 
\usepackage{amssymb}  
\usepackage{gensymb}

\usepackage{bm}
\usepackage{booktabs}

\title{\LARGE \bf
Strategic Jenga Play via Graph Based Dynamics Modeling
}

\author{Kavya Puthuveetil$^{1}$, Xinyi Zhang$^{2}$, Kazuto Yokoyama$^{2}$, and Tetsuya Narita$^{2}$
\thanks{$^{1}$Kavya Puthuveetil is with the Robotics Institute, Carnegie Mellon University, Pittsburgh, PA, USA.
        {\tt\footnotesize kavya@cmu.edu}}%
\thanks{$^{2}$Xinyi Zhang, Kazuto Yokoyama, and Tetsuya Narita are with Sony Group Corporation, Tokyo, Japan.
    {\tt\footnotesize tetsuya.a.narita@sony.com}}%
}

\begin{document}

\maketitle
\thispagestyle{empty}
\pagestyle{empty}

\begin{abstract}
Controlled manipulation of multiple objects whose dynamics are closely linked is a challenging problem within contact-rich manipulation, requiring an understanding of how the movement of one will impact the others. Using the Jenga game as a testbed to explore this problem, we graph-based modeling to tackle two different aspects of the task: 1) block selection and 2) block extraction. For block selection, we construct graphs of the Jenga tower and attempt to classify, based on the tower's structure, whether removing a given block will cause the tower to collapse. For block extraction, we train a dynamics model that predicts how all the blocks in the tower will move at each timestep in an extraction trajectory, which we then use in a sampling-based model predictive control loop to safely pull blocks out of the tower with a general-purpose parallel-jaw gripper. We train and evaluate our methods in simulation, demonstrating promising results towards block selection and block extraction on a challenging set of full-sized Jenga towers, even at advanced stages of the game.
\end{abstract}


\section{Introduction}


Contact-rich manipulation tasks, where the robot must manage contact between itself and its environment, remains a significant challenge that has been explored through a wide variety of tasks such as scooping, wiping, and peg-in-hole insertion~\cite{suomalainen2022survey}. Within contact-rich manipulation, multi-object manipulation is an important sub-skill where one must deal with collections of objects whose dynamics are interdependent and directly impacted by the robot's actions, as in tasks like object singulation, decluttering, or rearrangement~\cite{pan2022algorithms}. In these settings, handling one object can have stochastic effects on the other objects in the scene that can be irreversible or can otherwise cause task failure. Dealing with such scenarios requires reasoning about the paired dynamics of all the objects over some time horizon during control.

Jenga, a game that involves successively removing and replacing wooden blocks in an increasingly precarious tower, is one such task that requires joint reasoning about the collection of objects for successful manipulation. Small variations in the dimensions and weight of each block introduce different stresses in the tower, making some easy to remove, while the removal of others will cause the tower to collapse~\cite{marchionna2023deep}. Graphs are powerful tools for reasoning about complex relationships and physical interactions between objects, making graph-based learning approaches particularly well suited for multi-object manipulation tasks like Jenga, especially given the regular, underlying structure of the tower.

\begin{figure}[h]
    \centering
    \includegraphics[width=\linewidth, trim={17cm 7cm 0cm 5cm}, clip]{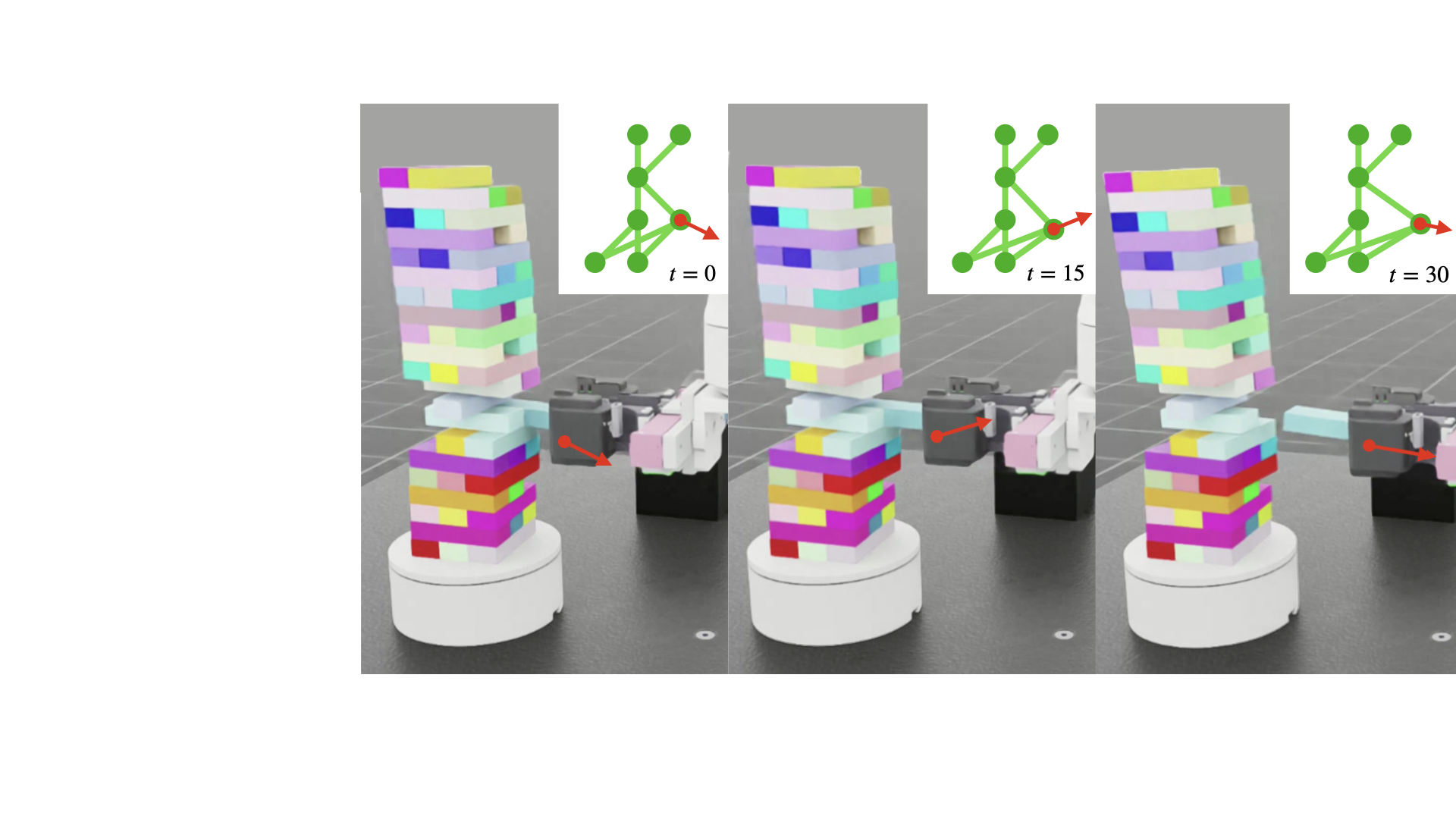}
    \caption{Successful extraction of a difficult-to-remove block from a Jenga tower on an advanced round (round nine) of the game. The coloration of the blocks in the tower is for visualization purposes only.}
    \label{fig:title_fig}
    \vspace{-0.6cm}
\end{figure}

In this work, we show how lightweight graphs can be used to model complex inter-block relationships and integrate high-level strategic modeling with low-level dynamics modeling and control. We consider the dynamics of the Jenga game in two different phases, block selection and block extraction, and present graph-based modeling techniques to reason about both. Block selection, the process of picking a block to remove from the tower, has been treated as a random or heuristic-based choice in several other works on robot Jenga play~\cite{fazeli2019seefeel, wang2009robot, marchionna2023deep, kroger2008manipulator}. However, humans play Jenga strategically, either picking blocks that are likely safe or adversarially choosing ones that make future extractions harder for their opponent. We attempt to imbue our robot with the foundational reasoning for strategic block selection by learning to classify whether a given Jenga tower, represented as a graph, will collapse if a specific block is removed.

For block extraction, we consider how to remove blocks from the tower, requiring small adjustments throughout the extraction to do so safely. We first learn a graph-based dynamics model of the tower that predicts, at a given timestep, how a proposed action to move one block will impact all its neighbors. We then use this learned dynamics model to produce samples for a Model Predictive Path Integral (MPPI) controller for the block extraction task~\cite{williams2017model, williams2018information, pezzato2025sampling}, allowing us to achieve higher control frequencies than if the simulator was used directly as the dynamics model. We train and evaluate our block selection and extraction methods in simulation on full-sized Jenga towers instantiated at different rounds of the game, in contrast to some previous works~\cite{wang2009robot, yoshikawa2011jenga}, showing robustness to a variety of difficult tower states.


Through this work, we make the following contributions:
\begin{itemize}
    \item We propose graph-based methods for reasoning about collections of objects simultaneously, explored via the strategic Jenga play task, which we split into block selection and block extraction phases
    \item We formulate block selection as a graph-based binary classification problem, learning whether removal of a specific block from a Jenga tower will cause it to fail.
    \item We demonstrate usage of a learned dynamics model in an MPPI controller to extract blocks in simulation.
    
\end{itemize}


\section{Strategic Jenga Play}
\begin{figure}
    \centering
    \includegraphics[width=\linewidth, trim={13cm 5cm 20cm 3cm}, clip]{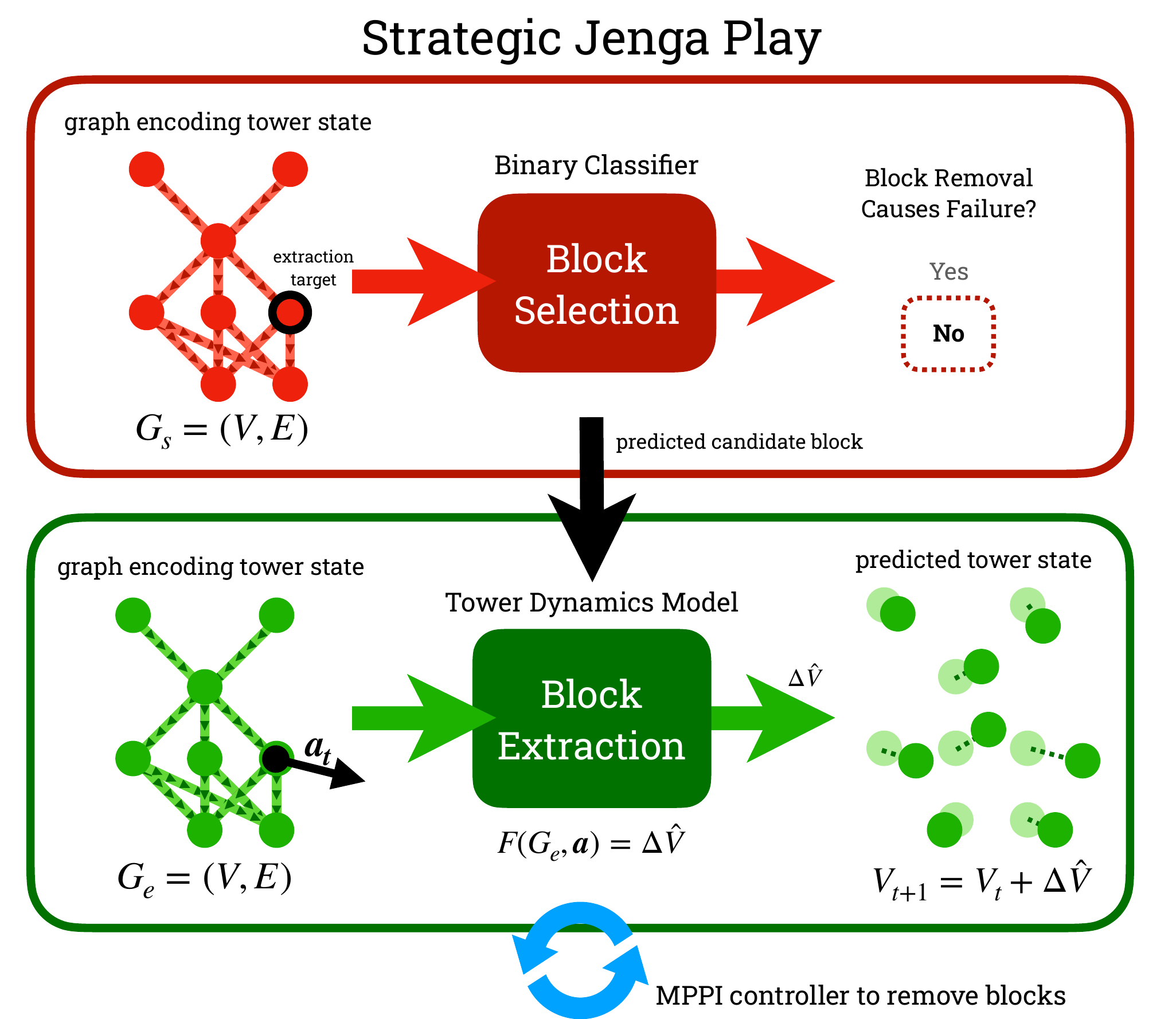}
    \caption{Overview of our graph-based approach to strategic Jenga play. For block selection, we train a classifier that predicts whether removing a given block from a tower, represented as a graph, will cause it to enter a failure state. For block selection, the tower state graph is fed into a dynamics model that predicts the displacement of the tower given an action that attempts to remove a candidate block. The block extraction dynamics model is used as part of an MPPI controller to carefully extract blocks.}
    \label{fig:overview}
    \vspace{-0.4cm}
\end{figure}

In this section, we describe our definition of the Jenga play task, simulation set-up, and task execution steps. Then, we detail our approach to reasoning about how to select blocks for removal, posing this as a classification problem where we seek to determine whether a given block can safely be extracted. Lastly, we describe our method for learning the dynamics of the Jenga tower during block extraction and establish an objective function that we apply in a sampling-based controller to remove blocks from the tower without causing it to collapse. We summarize our approach in Fig.~\ref{fig:overview}.

\subsection{Task Definition} 

\label{sec:task_definition}

We construct the Jenga task using the Isaac Lab framework ~\cite{mittal2023orbit}, a robot learning toolkit built on top of NVIDIA Isaac Sim. We initialize the simulation environment with a 7-DoF Franka robot arm outfitted with a general-purpose parallel jaw gripper (THK SEED+Picsel actuator with custom fingers), a turntable, a push-stick, and a full Jenga tower with 54 blocks (18 initial levels). We can randomly spawn Jenga towers between rounds 1-15, as shown in Appendix Fig.~\ref{fig:round_progression}, to ensure generalization of our methods across the spectrum of difficult tower states that appear as the game progresses. In the real world, the stresses in the tower that make block extraction challenging are caused in part by small variations in the weight and dimensions of the blocks. To replicate this variation, we randomly sample the 3D dimensions and weight of each block in the Jenga tower based on the distribution of measurements from a real Jenga set. We manually tuned the frictional parameters of the simulated blocks to match the properties of a real-world tower, where 47\% of blocks in a randomly initialized tower are movable, found empirically by Fazeli et al.~\cite{fazeli2019seefeel}.

\begin{figure}
    \centering
    \includegraphics[width=\linewidth, trim={2.5cm 1cm 5.2cm 1cm}, clip]{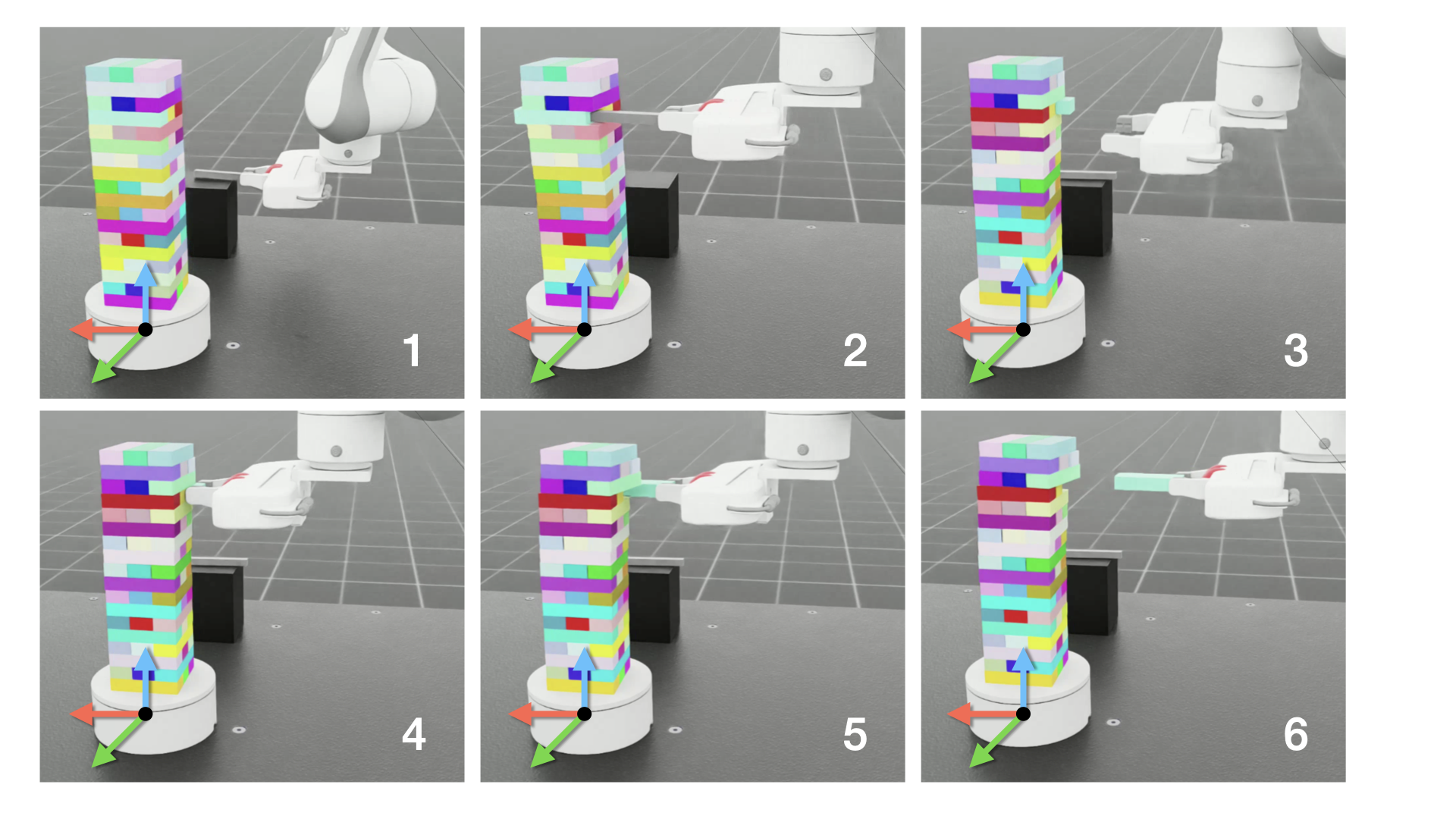}
    \caption{Key frames from a simulation rollout: 1) The tower is rotated so the target block is parallel to the $x$-axis and the robot picks up the push-stick, 2) the robot uses pushes the target block out of the tower slightly, 3) the tower rotates 180\degree and the robot drops the push stick, 4) the robot grasps the target block, 5) the robot begins pulling the block out in the $-x$-direction, and 6) the robot successfully extracts the block.}
    \label{fig:sim_rollout}
    \vspace{-0.6cm}
\end{figure}

The robot's interactions with the tower, shown in Fig.~\ref{fig:sim_rollout}, begin by selecting a given block to pull from the tower. To overcome limits in the robot's joint space and avoid the complex motion planning required to have the robot reach to all faces of the tower, we use a turntable, which the tower sits on, to rotate the tower such that the selected block is parallel to the robot's fingers and aligned to the $x$-axis. The robot picks up a push-stick to push the chosen block roughly 2 cm out of the tower, then puts down the push-stick while the turntable rotates 180\degree. All of the steps up to this point are executed using a state-machine, which we designed by hand. Block extraction is performed by a controller, which we introduce in Sec.~\ref{sec:block_extract}, and begins after the robot takes hold of the selected block, which now sticks out of the tower, in preparation to pull it out in the $-x$-direction. We define the block extraction action $\bm{a} = (a_x, a_y, a_z)$ as a 3D displacement in end effector space such that $a_y, a_z \in [-0.1, 0.1]$ cm.

\subsection{Block Selection}
\label{sec:block_select}
We define the Jenga block selection problem as a binary classification problem where, given the initial state of a Jenga tower and a candidate block for removal, we predict whether or not the selected block induces tower failure states. These failure states include: 1) a "stuck" extraction candidate block that cannot be displaced or 2) tower collapse.

We represent the initial state of the Jenga tower as a graph $G_s = (V, E)$ where the nodes $V = \{v_i\}_ {i=1...N}, N=54$ represent the blocks in the tower $J$ and edges $e_{ij} \in E$ represent the contact between blocks $v_i$ and $v_j$. For every block $j_i \in J$ in the tower, we create a node $v_i \in V$ that encodes the block's 3D dimensions, weight, and a 1-hot encoding of whether the block is a candidate to be removed. We create a directed edge, $e_{ij} \in E$, between a given block and the blocks immediately below with which it has contact. In Section~\ref{sec:block_select_results}, we compare this graph composition to an alternative, $G_{sp}$, that also encodes a given block's 7D pose in the nodes.

To perform graph-level classification of the constructed graphs for the block selection problem, we use a simple Graph Convolutional Network (GCN) with 6 convolutional layers. The training dataset for our GCN, containing 2,000 $(J_i, b, f)$ pairs, was collected from simulation rollouts where we try to extract a randomly selected block $b \in [1, N], N=54$ from initial tower state $J_i$ and observe whether the tower enters failure $f\in\{0, 1\}$. At each rollout, the Jenga tower is randomly spawned between rounds 1-15 and the target block to be removed is selected from a subset of all the blocks in the tower. This subset of blocks includes only valid extraction targets that could theoretically produce a stable tower (i.e. it does not include the center block in a level where there are only two blocks or the side blocks in a level that has only side blocks). Target blocks are extracted by applying a velocity to the selected block in the $-x$-direction with some Gaussian noise in the $y$, $z$ directions, and we observe whether the tower enters a failure state after the fact. We compose a graph for each of the $(J_i, b, f)$ pairs in the dataset and apply a 4:1 train-test split. The GCN is trained using the cross-entropy loss function over 150 epochs using the Adam optimizer, a learning rate of 1e-3, and a batch size of 100.

\subsection{Block Extraction}
\label{sec:block_extract}

Removing blocks from the tower is a dynamic control problem that requires forward reasoning about how a movement made now will influence the tower's stability in the distant future. We leverage the sampling-based Model Predictive Control method, Model Predictive Path Integral (MPPI)~\cite{williams2017model, williams2018information}, to design a reactive controller to extract Jenga blocks without disrupting the rest of the tower. Our implementation borrows from that presented by Pezzato and Salmi et al.~\cite{pezzato2025sampling} with one key difference: instead of directly using the simulator to generate future samples, we use a learned dynamics model. In this section, we describe our graph-based approach to modeling the Jenga tower's dynamics, as well as the objective function and parameters used in our MPPI control loop.

\subsubsection{Dynamics Modeling}
\label{sec:block_extract-dyn_modeling}
In order to learn the dynamics of the Jenga tower during block extraction, we use a similar graph representation of the tower as was described in Section~\ref{sec:block_select}. The key differences between the block selection graph $G_s$ and the block extraction graph $G_e$ are the node features. Specifically, for each node $v_i \in V$ in the extraction graph $G_e = (V, E)$, we continue to encode the corresponding block weight and a 1-hot encoding of the extraction target, as in $G_s$, but add the block's 7D pose and remove its 3D dimensions. We also treat the candidate action to extract a block from the tower, $\bm{a_t}$, as a global vector of the graph that is appended to each node.

The standard Graph Network-based Simulators (GNS) graph neural network architecture~\cite{sanchez-gonzalez2020learning, lin2021VCD} has been well demonstrated for effectively learning the dynamics of dependent particle systems like cloth. Since the dynamics of each block in the Jenga tower is closely influenced by it's neighbors, we choose to apply the GNS architecture to the Jenga dynamics learning problem. We build a GNS-based GNN with 4 processing layers that acts as a dynamics model $F(G_e, \bm{a_t}) = \Delta \hat{V}$ that, given a Jenga block extraction graph $G_e$ and a candidate action $\bm{a_t}$, outputs the predicted displacement of all nodes in the graph $\Delta \hat{V}$, or all blocks in the tower, after execution of the action. We can then compute the next state of the tower via $V_{t+1} = V_t + \Delta \hat{V}$.

In contrast to how we collected data to train the block selection model, we now use the robot to directly pull blocks out of the tower by executing an extraction trajectory over $T=51$ timesteps. The trajectory is made up of a series of actions, $\bm{a_1}, ..., \bm{a_T}$, which are defined as small displacements of the gripper in end effector space. We train our graph dynamics model on a dataset of 12000 $(J_t, b, \bm{a_t}, J_{t+1})$ pairs from random extraction interactions with the tower. At timestep $t$ in an extraction trajectory for a given simulation rollout, we observe how the initial state of the tower $J_t$ is impacted when block $b$ is pulled out of the tower by action $a_t$, resulting in a new tower state, $J_{t+1}$. Models are trained on the collected dataset over 150 epochs using the Adam optimizer with a mean squared error loss function, batch size of 100, and learning rate of 1e-4. Appendix Fig.~\ref{fig:gt_vs_pred} compares sample ground-truth final tower states $J_{t+1}$ and those predicted by our dynamics model $V_{t+1}$.

\subsubsection{MPPI for Block Extraction}

To optimize for actions that extract blocks without disrupting the rest of the tower, we define the following objective function:

$$
O(\bm{a_t}, b, V_t, V_{t+1}) = O_a(\bm{a_t}) + O_d(b, V_t, V_{t+1})
$$

The first term, $O_a(\bm{a_t}) = \alpha a_x + \beta |a_y| + \beta |a_z|$, attempts to both encourage movement in the primary extraction direction, the $-x$-direction, and discourage erroneous movements in the $y$ and $z$ directions. We use $\alpha=1000$ and $\beta=800$ as scalar weights to modify the respective rewards and penalties on the displacement of the end effector.

By contrast, the second term, 
$$O_d(b, V_t, V_{t+1}) = \gamma \sum^N_i \begin{cases} ||V_{t+1, i} - V_{t, i}|| & \text{if } i \neq b\\ 0 & \text{otherwise}\end{cases}$$
penalizes any displacement of blocks in the tower that are not the selected block $b$. We set the scalar weight $\gamma=100$, to balance this objective term relative to $O_a$.

Using our trained dynamics model and our objective function, we run our MPPI controller with a time horizon of 5 steps, performing 100 random samples at each step. We constrain the search space such that the maximum end effector displacement at a given timestep $t$ is $a_x, a_y, a_z \in [-2, 2]$cm, and set the coarseness of the search space to $\sigma$=1mm. While the computational expense of composing graphs to input into the dynamics model remains a rate-limiting factor, forward passes of the dynamics model to produce samples at each step of the time horizon are fast enough to maintain a control frequency of $\sim$30Hz, faster than what is achieved when using the simulator directly as the dynamics model~\cite{pezzato2025sampling}. In Sec.~\ref{sec:block_extract_results}, we compare the performance of using our full objective function $O(\bm{a_t}, b, V_t, V_{t+1})$ in the MPPI controller against just the first term $O_a(\bm{a_t})$, which acts as a naive baseline.

\section{Results and Discussion}
\label{sec:results&discussion}
\subsubsection{Block Selection}
\label{sec:block_select_results}
We evaluate our block selection GCN on 400 simulation rollouts held out from the training dataset. In Table~\ref{table:block_selection}, we present the classification accuracy of GCNs that are trained on two different graph representations of the tower, one that encodes the 7D pose of each block in the tower $G_{sp}$ and our primary method $G_s$, which does not.

\begin{table}
\centering
\caption{\label{table:block_selection} Block Selection Classification Results}
\begin{tabular}{cc} \toprule
    Graph Representation & Classification Accuracy\\ \midrule\midrule
    $G_{sp}$, Includes 7D Pose & 0.62\\
    $G_s$, Pose Not Encoded & $\bm{0.74}$ \\
	\bottomrule
\end{tabular}
\end{table}

\begin{table}
\centering
\caption{\label{table:block_extraction} Block Extraction Results}
\begin{tabular}{cc} \toprule
    Objective Function & Extraction Success Rate\\ \midrule\midrule
    $O_a(\bm{a_t})$ & 0.54\\
    $O(\bm{a_t}, b, V_t, V_{t+1})$& $\bm{0.65}$ \\
	\bottomrule
\end{tabular}
\vspace{-0.5cm}
\end{table}

Our method successfully predicts whether removing a given block causes a failure state of the tower 74\% of the time on the test set, which contains a challenging distribution of tower initializations between rounds 1-15 of the game. Including the 7D pose of each block in the node features of the graph has no benefit on performance, demonstrating that localization of the accurate, absolute pose of each block, a significant challenge in the real world, is not necessary to reason about block selection - the relative arrangement of the blocks, as well as their weight and dimensions, are enough.

\subsubsection{Block Extraction}
\label{sec:block_extract_results}
To evaluate our learned dynamics model and MPPI controller for block extraction, we attempt to remove randomly selected blocks in 100 simulation rollouts and record the success rates in removing those blocks without causing the tower to collapse, as summarized in Table~\ref{table:block_extraction}. Our controller, with cost function terms that both constrain the displacements of the robot's end effector and the displacements of the blocks in the tower, achieves a success rate of 65\%. Compared to our naive baseline, which uses only the $O_a(\bm{a_t})$ objective term, the success rate drops to 54\%, demonstrating that a naive controller that does not explicitly reason about how removing the target block influences the tower's dynamics is insufficient.

\begin{figure}
    \centering
    \includegraphics[width=\linewidth, trim={10cm 7.6cm 13cm 10cm}, clip]{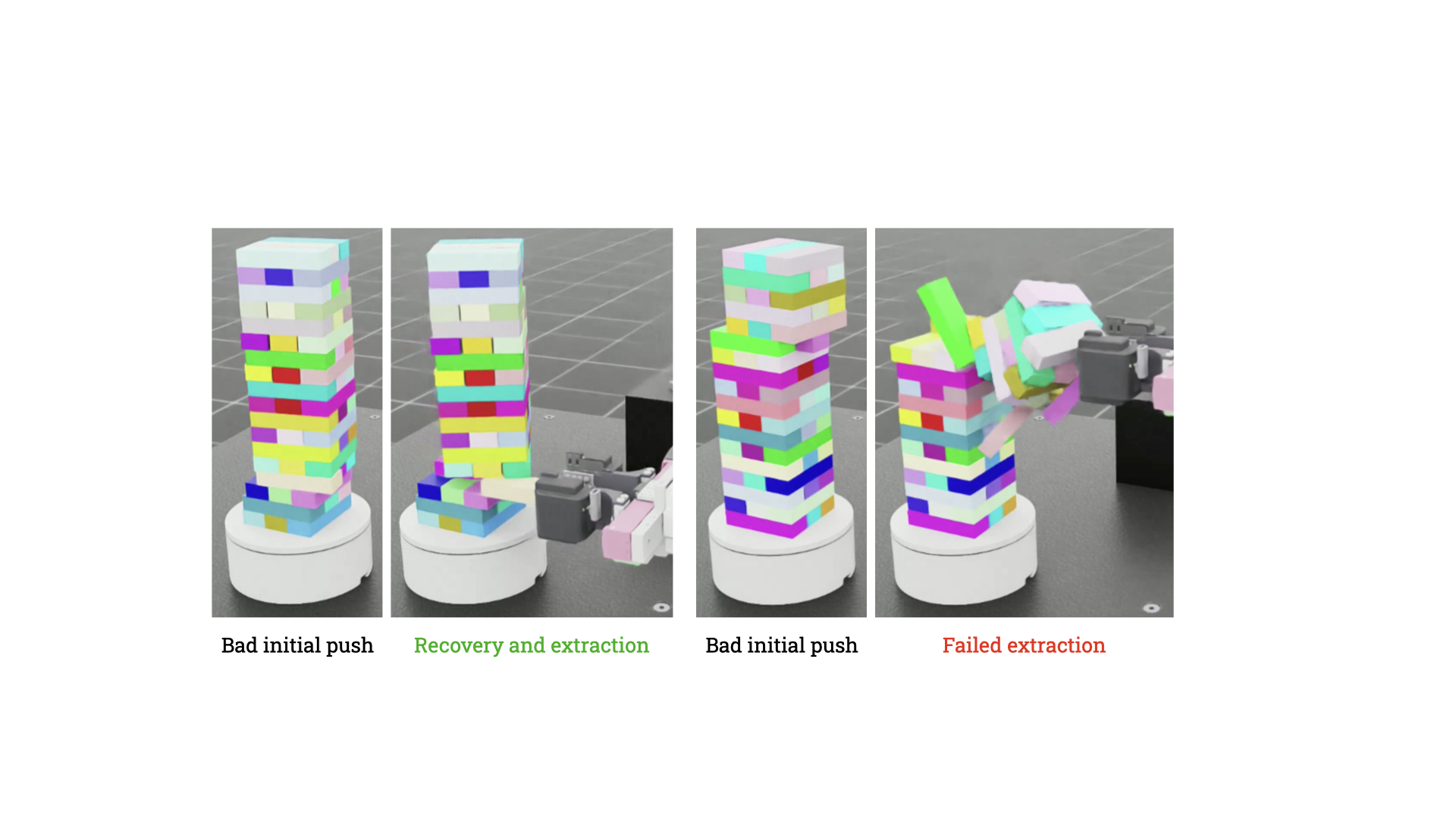}
    \caption{Example recovery and failure cases with bad initial pushes.}
    \label{fig:bad_push_cases}
    \vspace{-0.6cm}
\end{figure}

Among the failures we observed, many were caused by the tower being in an initial configuration that inhibited extraction. Specifically, the initial push of a target block, executed naively using our hand-defined state-machine, sometimes displaces entire layers of the tower instead of a single block, as shown in Fig.~\ref{fig:bad_push_cases}. Thus, in the extraction phase, handled by our controller, the robot struggles to grasp only the target block or to pull the target block out of the tower without displacing adjacent layers. This scenario often occurred when the randomly selected target block was "stuck" in the tower from the beginning, indicating that future integration of the block selection and block extraction modules could help to mitigate this type of failure. Applying a dynamic controller to perform the initial push, as well as the full extraction, could further prevent this behavior.

\subsubsection{Future Work}
\label{sec:future_work}
Our current implementation of block extraction, using the learned graph-based dynamics model in the MPPI control loop, already achieves higher control frequencies than when the simulator is directly used for the same~\cite{pezzato2025sampling}. However, we are currently limited by the graph composition phase, a relatively expensive process of creating a graph for each sample computed at each timestep in the time horizon for MPPI. Since running forward passes of our dynamics model has near trivial time cost, speeding up the graph composition on GPU could enable us to achieve even higher rates of control.

In Appendix Fig.~\ref{fig:extract_contact}, we present 3-axis contact measurements at the robot's fingertips captured during both successful and unsuccessful representative extraction attempts. As one would intuitively expect, the contact values are significantly more erratic in the failed extraction case and smoother in the successful extraction, signaling the potential utility of these signals in control. In future work, we hope to leverage these contact values to improve extraction accuracy and to help overcome potential challenges in transferring our method to the real world. For example, our block extraction dynamics model relies on the 7D pose of the blocks in the tower, which is non-trivial to reconstruct in the real world; fusion of vision and tactile methods may reduce the need for such accurate pose reconstruction~\cite{fazeli2019seefeel, marchionna2023deep}.

\section{Conclusion}
In this work, we explore how graph-based methods can be used to reason about manipulating groups of interdependent objects through the strategic Jenga play task. Our block selection GCN allows us to predict whether a given block is integral to the tower structure, enabling more strategic selection of what blocks to remove. Learning a dynamics model of the tower, which is then used in an MPPI control loop for block extraction, allows us to carefully pull blocks out of challenging tower configurations at relatively high control speeds. Our simulation results are promising for future work on sim-to-real transfer of our methods, as well as visual-tactile fusion for greater robustness in the real world.





\subsection{Appendix}

\begin{figure}[h]
    \centering
    \includegraphics[width=0.9\linewidth, trim={8.5cm 4.6cm 7cm 7cm}, clip]{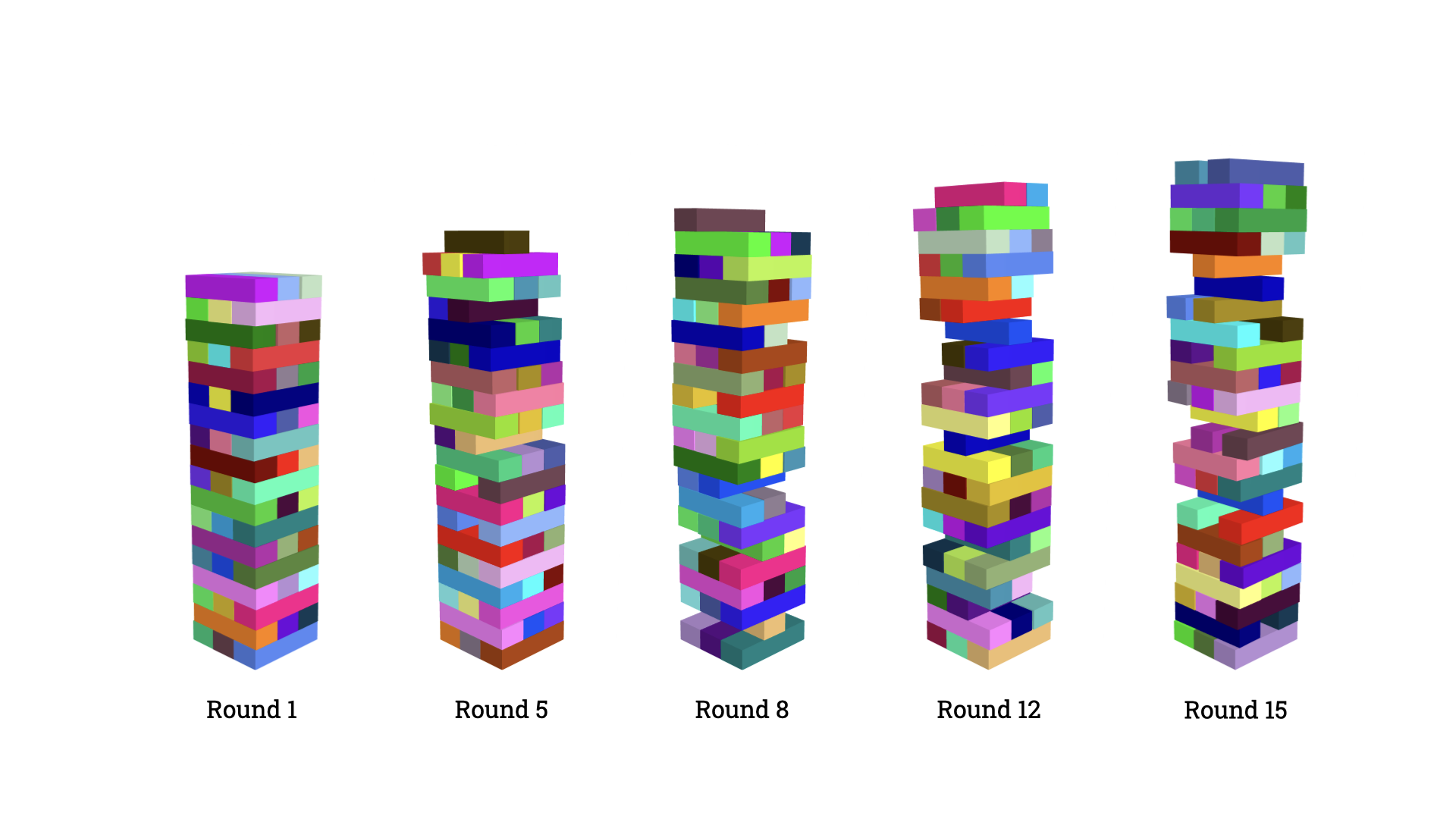}
    \caption{Sample tower initializations at different rounds of the game. The coloration of the blocks in the tower is for visualization purposes only.}
    \label{fig:round_progression}
\end{figure}

Appendix Fig.~\ref{fig:round_progression}, which showcases a set of sample towers between rounds 1-15 of the Jenga game, is discussed in Sec.~\ref{sec:task_definition}.

\begin{figure}[h]
    \centering
    \includegraphics[width=\linewidth, trim={5.5cm 8cm 8cm 3cm}, clip]{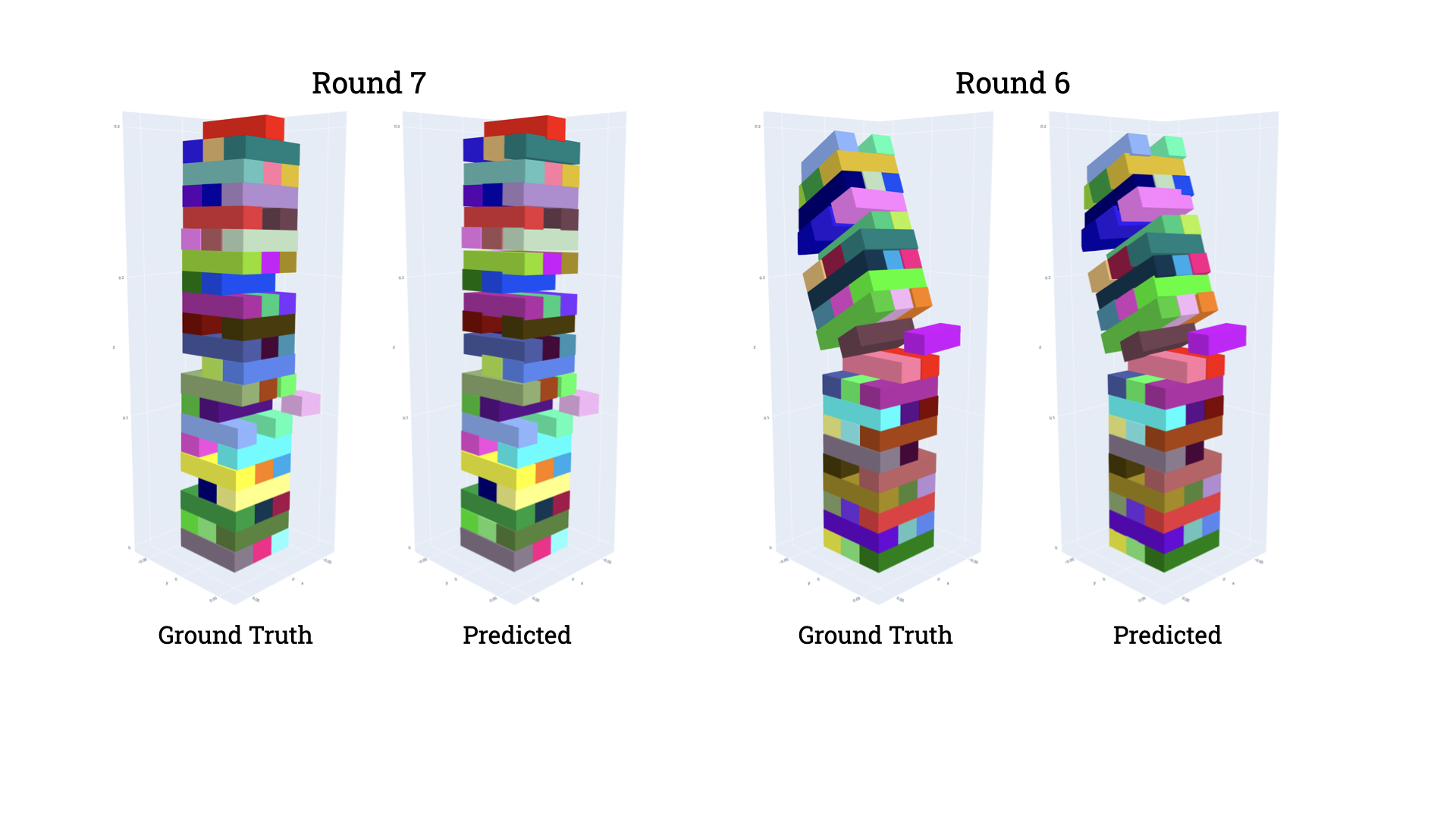}
    \caption{Examples of final tower states predicted by our dynamics model, given an input graph with a candidate extraction action, compared to the corresponding ground truth tower state, which our predictions closely match.}
    \label{fig:gt_vs_pred}
\end{figure}

Appendix Fig.~\ref{fig:gt_vs_pred} shows examples of comparisons between the ground truth tower state produced by a given extraction action and the corresponding predicted tower state from our dynamics model. This figure is discussed in Sec.~\ref{sec:block_extract-dyn_modeling}.

\begin{figure}[h]
    \centering
    \includegraphics[width=\linewidth, trim={12.5cm 0cm 10.5cm 0cm}, clip]{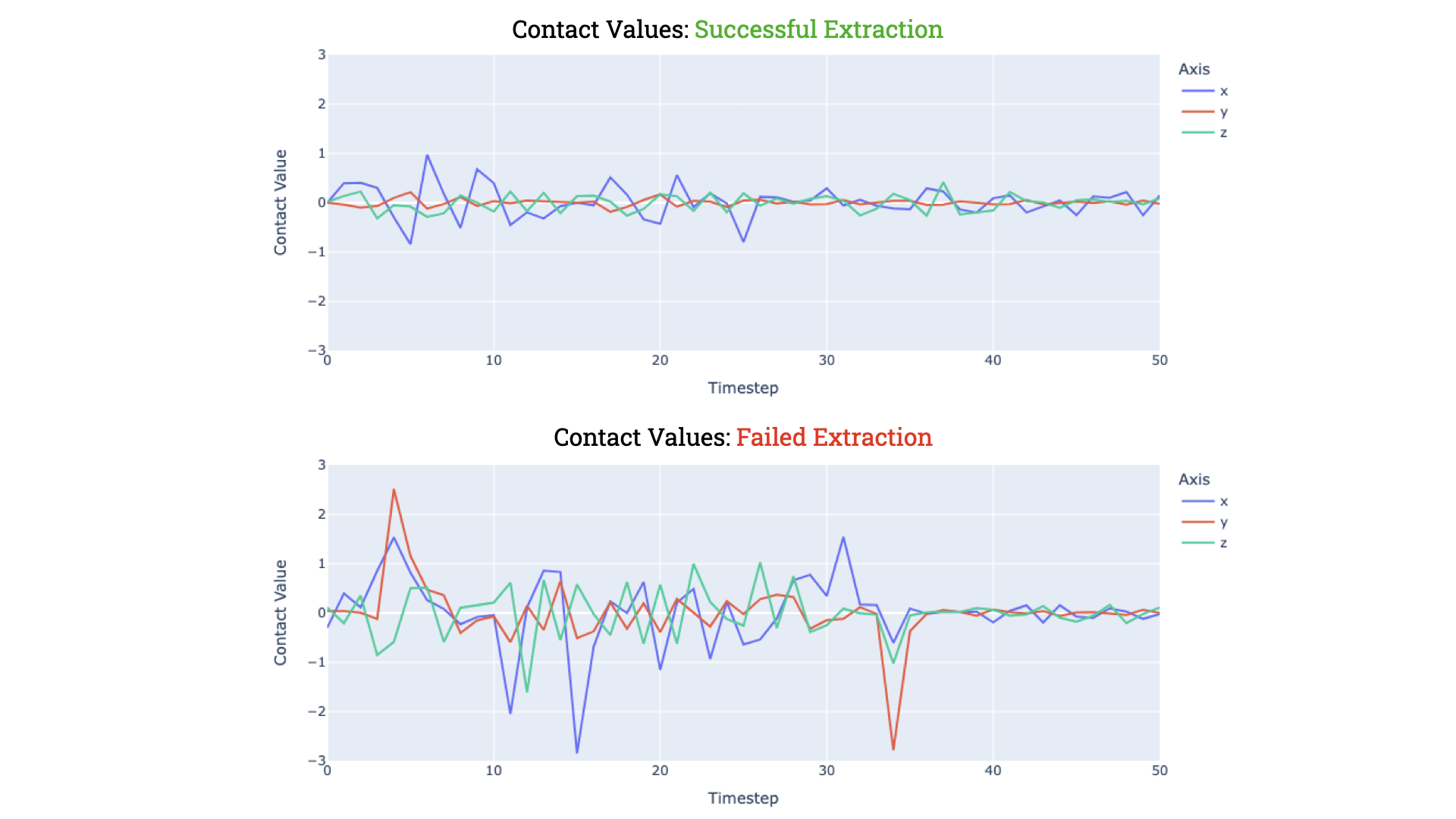}
    \caption{Sample contact values in the $x$, $y$, and $z$-axes measured during both successful and failed extractions. Values are far more erratic in cases where the extraction is unsuccessful.}
    \label{fig:extract_contact}
\end{figure}

Appendix Fig.~\ref{fig:extract_contact}, which shows contact values at the robot's end effector during successful and failed extractions, is mentioned in Sec.~\ref{sec:future_work}.

\section*{Acknowledgment}
We extend our gratitude to Takahisa Ueno and Daisuke Yamada for their valuable comments and insights on how to formulate our approach with consideration towards real-world implementation.



\bibliographystyle{IEEEtran}
\bibliography{root}

\end{document}